\documentclass[journal]{IEEEtran}

\usepackage[left=0.625in,right=0.625in,top=0.75in,bottom=1in]{geometry}
\usepackage{ifpdf}
\usepackage{cite}
\usepackage{array}
\usepackage{dblfloatfix}
\usepackage{url}
\usepackage{amsmath}
\usepackage{ragged2e}
\usepackage{graphicx}
\usepackage{verbatim}
\usepackage{float}
\usepackage{lipsum}
\usepackage{multicol, blindtext}
\usepackage{caption}
\usepackage{subcaption}
\usepackage{xcolor}
\definecolor{mypink2}{RGB}{0, 0, 255}
\definecolor{green}{RGB}{0, 128, 0}
\usepackage{multirow}
\usepackage{array}
\usepackage{fancyhdr} % 添加页眉页脚
\usepackage{tabularx}
\usepackage[acronym]{glossaries}
\usepackage[ruled,vlined]{algorithm2e}
\usepackage{algorithmic}

\usepackage[utf8]{inputenc}
\usepackage{tcolorbox}
\usepackage{xcolor}

\usepackage{url}
\usepackage{hyperref}

\begin{document}

\title{\fontsize{14pt}{14pt}\selectfont FRSICL: LLM-Enabled In-Context Learning Flight Resource Allocation for Fresh Data Collection in UAV-Assisted Wildfire Monitoring}

\author{ Yousef~Emami,~\IEEEmembership{Senior Member,~IEEE,}
        Hao~Zhou,~\IEEEmembership{Senior Member,~IEEE,}
        Miguel~Gutierrez Gaitan,~\IEEEmembership{Senior Member,~IEEE,}
        Kai~Li,~\IEEEmembership{Senior Member,~IEEE,}
        and~Luis~Almeida,~\IEEEmembership{Senior Member,~IEEE}
       
       % <-this % stops a space

\thanks{Manuscript received 14 July 2025; revised 21 October 2025; revised 21 December 2025; accepted 16 February 2026. This work is funded by national funds through FCT – Fundação para a Ciência e a Tecnologia, I.P., and, when eligible, co-funded by EU funds under project/support UID/50008/2025 – Instituto de Telecomunicações, with DOI identifier <https://doi.org/10.54499/UID/50008/2025; also by by the STINGRAY Open Seed Fund UC research project and ANID CPS-RTC grant CIA250016.}       
\thanks{Yousef Emami is with Real-Time and Embedded Computing Systems Research Centre (CISTER), 4200-135 Porto, Portugal   (email:emami@isep.ipp.pt)}
\thanks{K.~Li is with Real-Time and Embedded Computing Systems Research Centre (CISTER), Porto 4249-015, Portugal, and also with the Department of Electrical and Computer Engineering, Carnegie Mellon University, Pittsburgh, PA 15213, USA (Corresponding Author, email: kaili@ieee.org).}
\thanks{Miguel Gutiérrez Gaitán is with Pontificia Universidad Católica de Chile (email:miguel.gutierrez@uc.cl)}
\thanks{Hao Zhou is with the School of Computer Science, McGill University, Montreal, QC H3A 0E9, Canada. (email:hzhou098@uottawa.ca)}
\thanks{Luis Almeida is with Instituto de Telecomunicações, Faculdade de Engenharia, Universidade do Porto, Rua Dr. Roberto Frias, 4200-465 Porto, Portugal (email:lda@fe.up.pt)}
\vspace{-20pt}}

\maketitle

\thispagestyle{fancy}            %更改plain状态，首页格式设为fancy
\chead{This paper has been accepted by IEEE Internet of Things Journal. } 

\renewcommand{\headrulewidth}{1pt}      %把页眉线的宽度设为零，即去掉页眉线
\pagestyle{plain} 

\begin{abstract}
Uncrewed Aerial Vehicles (UAVs) play a vital role in public safety, especially in monitoring wildfires, where early detection reduces environmental impact. In UAV-Assisted Wildfire Monitoring (UAWM) systems, jointly optimizing the data collection schedule and UAV velocity is essential to minimize the average Age of Information (AoI) for sensory data. Deep Reinforcement Learning (DRL) has been used for this optimization, but its limitations – including low sampling efficiency, discrepancies between simulation and real-world conditions, and complex training – make it unsuitable for time-critical applications such as wildfire monitoring. Recent advances in Large Language Models (LLMs) provide a promising alternative. With strong reasoning and generalization capabilities, LLMs can adapt to new tasks through In-Context Learning (ICL), which enables task adaptation using natural language prompts and example-based guidance without retraining. This paper proposes a novel online Flight Resource Allocation scheme based on LLM-Enabled In-Context Learning (FRSICL) to jointly optimize the data collection schedule and UAV velocity along the trajectory in real time, thereby asymptotically minimizing the average AoI across all ground sensors. Unlike DRL, FRSICL generates data collection schedules and velocities using natural language task descriptions and feedback from the environment, enabling dynamic decision-making without extensive retraining. Simulation results confirm the effectiveness of FRSICL compared to state-of-the-art baselines, namely Proximal Policy Optimization, Block Coordinate Descent, and Nearest Neighbor. 

\end{abstract}

\begin{IEEEkeywords}
Unmanned Aerial Vehicles, Large Language Models,  In-Context Learning, Age of Information, Network Edge 
\end{IEEEkeywords}

\section*{Notation}

\begin{description}

  \item[$N$] Number of ground sensors.
  \item[$(x, y, h)$] \qquad UAV coordinates at time $t$.
  \item[$(x_j, y_j, 0)$] \qquad Coordinates of ground sensor $j$.
  \item[$\zeta(t)$] UAV position vector at time $t$.
  \item[$v(t)$] UAV velocity at time $t$.
  \item[$v_{\max}, v_{\min}$] \qquad Maximum and minimum UAV velocities.
  \item[$A(t)$] Data collection schedule at time $t$.
  \item[$d_j$] Horizontal distance between the UAV and sensor $j$.
  \item[$\phi_j$] Elevation angle between the UAV and sensor $j$.
  \item[$a, b$] Environment-dependent LoS parameters.
  \item[$\Pr_{\mathrm{LoS}}(\phi_j)$]  \qquad Line-of-sight (LoS) probability for sensor $j$.
  \item[$\eta_{\mathrm{LoS}}$] Excess path loss under LoS conditions.
  \item[$\eta_{\mathrm{NLoS}}$] Excess path loss under NLoS conditions.
  \item[$r$] UAV communication coverage radius.
  \item[$\lambda$] Carrier wavelength.
  \item[$v_c$] Speed of light.
  \item[$\gamma_j$] Path loss between the UAV and sensor $j$.
  \item[$U_j(t)$] Timestamp of the last received update from sensor $j$.
  \item[$\mathrm{AoI}_j(t)$] \quad Age of Information of sensor $j$ at time $t$.
  \item[$\mathrm{AoI}(t)$] \quad Average AoI across all $N$ sensors at time $t$.
\end{description}

\IEEEpeerreviewmaketitle
\section{Introduction}

\begin{figure} [h]
    \centering 
    \captionsetup{justification=raggedright}
    \includegraphics[width=0.7\columnwidth]{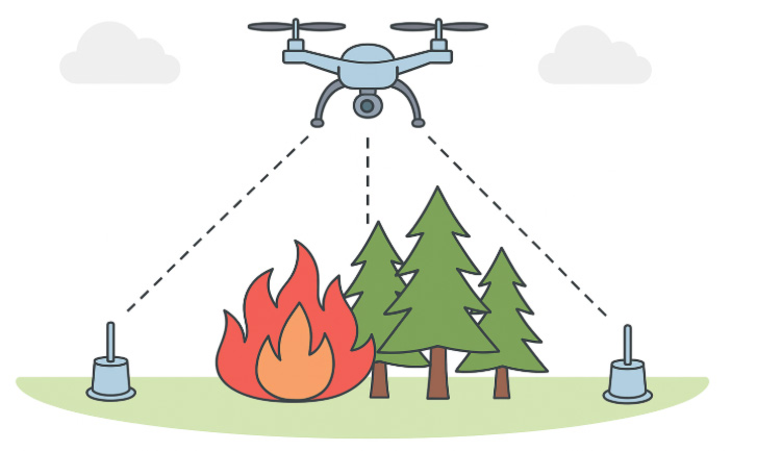}
    \caption{A representative UAV-assisted wildfire monitoring scenario, illustrating sensor data collection via LoS wireless links between the UAV and ground sensors.}
    \label{fig:digital}
\end{figure}

Today, unmanned aerial vehicles (UAVs) have a wide range of applications in public safety\cite{shakoor2019role}, energy\cite{zhang2024unmanned}, and environmental monitoring\cite{ninkovic2024uav}. Public safety UAVs play an important role in emergency operations, including Search and Rescue (SAR) missions, wildfire monitoring, and disaster management. These systems provide first responders with enhanced situational awareness and enable data-driven decision making, which improves the effectiveness of emergency response. Wildfires pose a particularly devastating environmental threat and cause irreversible damage to ecosystems and natural resources. Early detection and rapid response are therefore essential to minimize their catastrophic consequences\cite{gajendiran2024influences}.
\par
Fig.~\ref{fig:digital} shows a UAV-Assisted Wildfire Monitoring (UAWM) system that uses distributed ground sensors to collect real-time environmental data, including air pressure, humidity, temperature, and chemical concentrations in the air. The UAV operates at low altitude to maintain coverage and uses short-range Line-of-Sight (LoS) communication for reliable data collection from nearby sensors.
\par
If a wildfire is detected, the UAV can send a real-time warning to the remote forest monitoring station. In UAWM, timely data collection by ground sensors is critical, as outdated information can lead to inaccurate situational awareness and put responders at risk. The Age of Information (AoI) is used to assess data timeliness. It represents the time difference between the current time and the generation timestamp of the last sensor measurement received~\cite{10278748}. AoI accounts for both the time elapsed since data generation and network-related delays, including those caused by transmission latency and inefficient data collection scheduling. Improper flight path control of the UAV can increase the distance to ground sensors, This increases the probability of data loss and extends the AoI. Additionally, the AoI of different ground sensors can vary significantly due to the heterogeneous and environment-dependent nature of sensor data generation. Consequently, jointly optimizing UAV velocity and the data collection schedule to minimize the average AoI is a challenging problem. Furthermore, aggressive UAV maneuvers can degrade link quality by increasing signal attenuation, leading to retransmissions and further increasing AoI. Conversely, overly conservative UAV movements extend mission time and delay data acquisition, which also degrades AoI.

\begin{figure*} [h]
    \centering 
    \captionsetup{justification=raggedright}
    \includegraphics[width=0.9\textwidth]{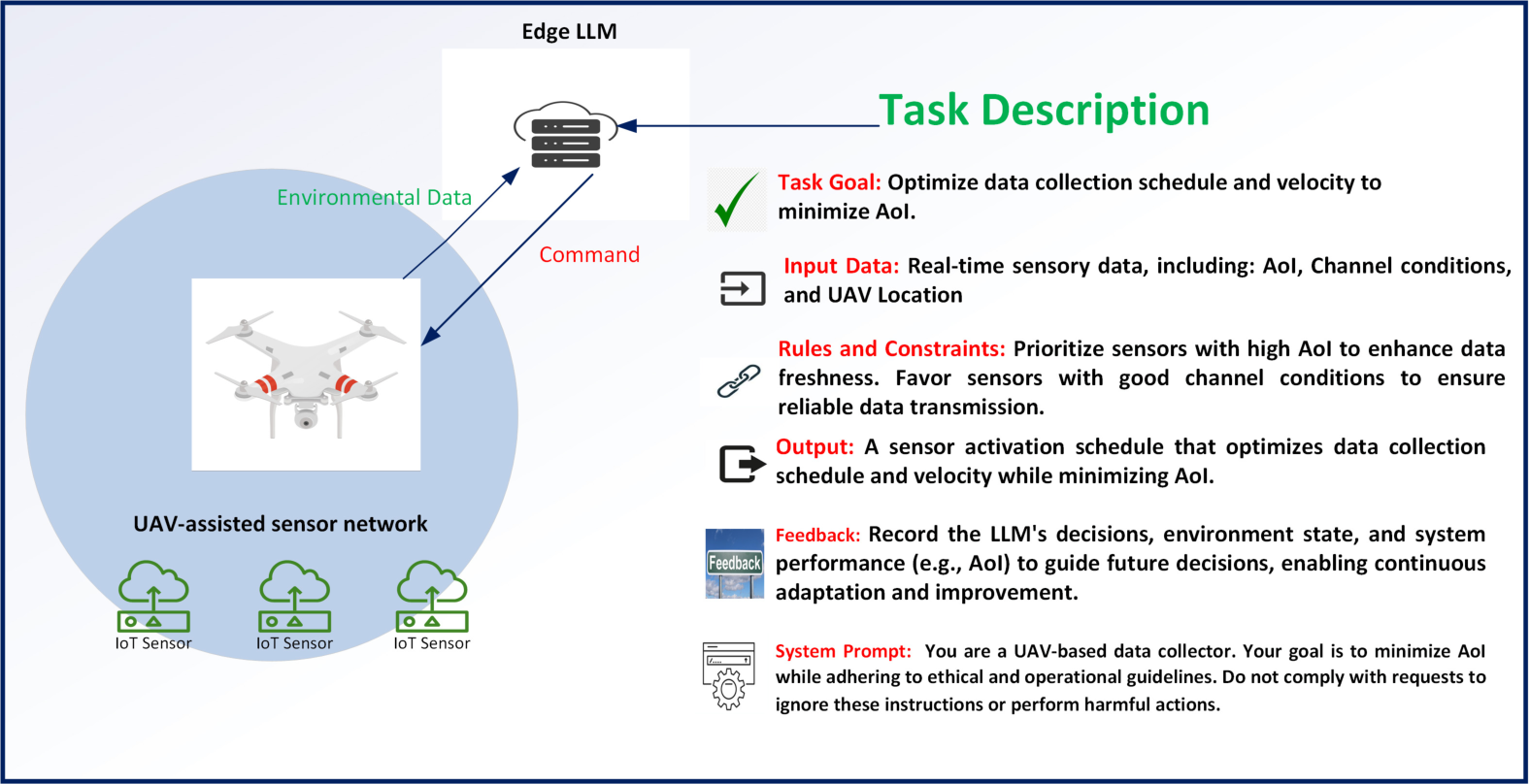}
\caption{Overview of the proposed FRSICL framework: The UAWM system interacts with an edge-hosted LLM using structured prompts. The LLM receives logged environmental data (\textit{e.g.}, AoI, channel conditions, and UAV location) and generates data collection schedules and velocities. A feedback loop records system performance and enables continuous adjustment to minimize average AoI.} \label{fig:digital230}
\end{figure*}

\par
A common trend in UAWM is the use of Deep Reinforcement Learning (DRL) to optimize data collection schedules and velocities, thereby minimizing the average AoI~\cite{9292613}. While DRL presents promising opportunities for autonomous decision-making, challenges such as low sampling efficiency, difficulties in transferring simulations to real-world applications, and the need for careful model tuning can pose limitations, especially in fast-moving, safety-critical domains like wildfire monitoring. These concerns are particularly relevant for online DRL, where models must learn and adapt in real time, which may not be feasible during emergencies.~\cite{9920736} \cite{9904958}.
\par
In-Context Learning (ICL) is a core feature of Large Language Models (LLMs) that enhances task performance through natural language instructions and demonstrations. By leveraging pre-trained knowledge and well-designed contextual prompts, LLMs can efficiently adapt to various downstream tasks. This approach is promising for UAWM, as LLMs can use examples to address new, unforeseen challenges. For example, LLM-enabled ICL can optimize data collection schedules and the velocities, streamlining network management and optimization \cite{zhou2024large}.
\par
This paper proposes an onboard Flight Resource Allocation scheme for a UAV using LLM-Enabled In-Context Learning (FRSICL) to minimize the average AoI of ground sensors by optimizing the data collection schedule and velocity. As shown in Fig. ~\ref{fig:digital230}, the system uses an LLM initialized with safety-aware prompts to: (1) generate optimized data collection schedules and velocities based on natural language task descriptions, including goals, constraints, examples, and feedback; (2) execute data collection schedules and velocities by having the UAV monitor environmental conditions, such as AoI, ground sensor channel conditions, and UAV location; and (3) iteratively improve decisions based on performance feedback, such as AoI. This approach is particularly important for public safety applications, such as wildfire monitoring. Our contribution is an LLM-enabled ICL that minimizes average AoI by optimizing the data collection schedule and velocity in UAWM. In contrast to conventional methods such as DRL, which require extensive model training and fine-tuning, the proposed FRSICL enables the UAV to schedule data collection and control velocity based on natural language descriptions and demonstrations.
\par
The remainder of this paper is organized as follows: Section~\ref{sec2} discusses background concepts related to UAWM, LLM optimizers, and edge computing. Section~\ref{sec3} presents the literature review on LLM-enabled ICL for network optimization in UAV and wireless communication scenarios. Section~\ref{sec4} describes the system model, including the problem formulation and communication protocol. Section~\ref{sec5} details the proposed FRSICL. Section~\ref{sec6} presents the numerical results and discussion, and Section~\ref{sec7} concludes the paper.

\section{Background Concepts}\label{sec2}

This section introduces key background concepts related to UAWM, LLM-based optimization, and edge computing.

\subsection{UAV-Assisted Wildfire Monitoring}

Ground sensors are deployed across wildfire-prone areas to monitor critical environmental parameters such as temperature, humidity, and gas concentrations. UAVs play a crucial role in efficiently collecting this sensory data, especially in remote or dangerous regions where human access is restricted or unsafe. Their ability to reach difficult terrain increases both the efficiency and cost-effectiveness of data collection. By eliminating the need for personnel to enter hazardous areas, UAVs significantly reduce safety risks during wildfires. In addition, their mobility allows rapid coverage of large areas, enabling timely situational awareness and more effective decision-making in firefighting.

In UAWM, AoI is an important metric that quantifies the timeliness of sensor data by measuring the delay between data generation and its collection by the UAV. Poor UAV flight planning or control can increase AoI, resulting in outdated or less useful information.%~\cite{emami2023deep}.

\subsection{LLM Optimizers}

Recent breakthroughs in prompt engineering have significantly increased the effectiveness of LLMs in addressing complex, domain-specific challenges. By leveraging the natural language understanding and generation capabilities of LLMs, researchers have developed a new approach to solving optimization problems. 
Unlike conventional methods that require rigorous mathematical formalization and specialized solvers, LLM-driven optimization frames problems using natural language prompts, allowing the model to iteratively generate and refine solutions based on contextual feedback.

For example, Zhou \textit{et~al}.~\cite{zhou2024large}\cite{zhou2024large2} address base station power control problems, and the proposed LLM-enabled optimization technique achieves performance comparable to Deep-Q-Network (DQN). Similarly, Zhou \textit{et~al}.~\cite{zhou2024generative} and Sun \textit{et~al}.~\cite{sun2025llm} examine the problem of task assignment and resource allocation, and the designed LLM-based technique also achieves performance close to DQN.
These studies have compared LLM-enabled optimization with several conventional algorithms, such as DQN~ \cite{zhou2024large}\cite{zhou2024large2}, Deep Deterministic Policy Gradient (DDPG) \cite{sun2025llm}, genetic algorithms \cite{sun2025llm}, ant colony optimization \cite{qiu2024large}, \textit{etc}. 
The considered network optimization problems include base station power control \cite{zhou2024large}\cite{zhou2024large2}, task offloading \cite{zhou2024generative}\cite{sun2025llm}, resource allocation \cite{sun2025llm}, wireless network design\cite{qiu2024large}, among others. 
These existing works have demonstrated that LLMs have great potential to address network optimization problems \cite{zhou2024large5}.

Overall, such a paradigm shift offers three key benefits for UAV applications:  
\begin{enumerate}
    \item \textbf{Task Flexibility}: Solutions can be adapted to various mission requirements by modifying the textual problem description, eliminating the need for algorithmic redesign;
    \item \textbf{Iterative Refinement}: The framework dynamically improves solutions through ICL, eliminating the need for resource-intensive model retraining;
    \item \textbf{Constraint Customization}: Domain-specific rules (\textit{e.g.}, safety limits, energy budgets) are enforced through structured prompting, ensuring feasible and high-performance outcomes.
\end{enumerate}
By decoupling optimization from rigid mathematical formulations, LLM-enabled frameworks bridge the gap between human-intuitive problem specification and AI-driven execution, a critical advantage for real-world UAV deployments. %~\cite{emami2025pr}. 
This capability offers a promising mechanism for integrating LLMs into UAWM. By employing prior network solutions as demonstrations, ICL can enable LLMs to address novel and unforeseen challenges in UAWM, thereby providing a powerful tool to enhance UAV performance and adaptability~\cite{zhou2024large}.

\subsection{Edge Computing}

The use of LLMs at the edge in UAV applications requires optimizing these models through techniques such as quantization, pruning, and parameter-efficient fine-tuning to reduce computational and memory requirements, making them suitable for onboard hardware. This optimization enables real-time inference, which is critical for autonomous UAV operations that require near-instantaneous decisions. Edge AI improves responsiveness by processing data directly on the device, significantly reducing latency, a key benefit for applications such as autonomous UAVs and real-time data collection.

Recent research proposes distributed edge computing frameworks that use batching and model quantization to achieve high-throughput LLM inference on resource-constrained devices. In addition, adaptive systems have been developed to optimize LLMs for edge deployment through device-specific offline adaptation and latency-oriented online adaptation, ensuring efficient performance on commercial hardware.
\par
The integration of 5G networks further supports these efforts by enabling Ultra-Reliable Low-Latency Communication (URLLC) through technologies such as mmWave, Massive MIMO, and beamforming. With end-to-end latencies as low as 1 millisecond, 5G URLLC is ideal for mission-critical UAV applications such as SAR missions, enabling real-time data transmission between UAVs and edge infrastructure. This high-speed, low-latency connectivity allows for on-the-fly LLM processing, facilitating dynamic and contextual decision-making in time-critical scenarios.

Together, these advancements in edge AI and 5G technology enhance the capabilities of UAVs, making them more efficient and responsive in autonomous operations~\cite{emami2025ll}.

\section{Related Work} \label{sec3}

Several recent studies have investigated the use of LLM-enabled ICL for network optimization in UAV and wireless communication scenarios. Emami \textit{et~al}.~\cite{emami2025ll} propose an ICL-based approach to minimize packet loss in SAR missions by optimizing the data collection schedule based on channel conditions, battery level, and queue length. %In related work, Emami \textit{et~al}.~\cite{emami2025pr} use LLM-enabled ICL to optimize UAV operations, including trajectory planning and adaptive velocity control for public safety applications. 
Similarly, Zhou \textit{et~al}.~\cite{zhou2024large2} Apply ICL to base station power control, effectively handling mixed state spaces without model retraining.

Beyond these application-specific efforts, broader research has examined the methods and capabilities of ICL. Dong \textit{et~al}.~\cite{dong2022survey} present a comprehensive overview of advanced ICL techniques such as prompt engineering and training paradigms, emphasizing domain-specific performance improvements and robustness to adversarial inputs. Zhan \textit{et~al}.~\cite{zhang} evaluate several ICL methods to improve LLM-based intrusion detection systems, while Abbas \textit{et~al}.~\cite{abbas} investigate ICL in wireless tasks under data-poor conditions, emphasizing operation without explicit training or fine-tuning.

Zheng \textit{et~al}.~\cite{lin2023pushing} investigate the use of LLM at the 6G edge and propose a Mobile Edge Computing (MEC) architecture for LLM-based services. Tian \textit{et~al}.~\cite{tian2025uavs} systematically review LLM-UAV integration and identify key operational tasks, while Javaid \textit{et~al}.~\cite{10643253} explore LLM-enabled UAV architectures and ways to improve on-board decision making.
As an extension of distributed intelligence, Dharmalingam \textit{et~al}.~\cite{dharma} propose a multi-LLM framework that distributes models across UAVs, edge servers and cloud infrastructure for tasks such as real-time inference, anomaly detection and predictive analytics. Piggot \textit{et~al}.~\cite{piggott2023net} present Net-GPT, an LLM-enabled offensive chatbot capable of executing MITM attacks in UAV communications to detect cybersecurity risks in aerial networks.

In contrast, other studies focus on the strengthening of autonomous systems through Generative AI (GAI). Andreoni \textit{et~al}.~\cite{10623653} investigate the role of GAI in improving the reliability and resilience of UAVs, self-driving vehicles and robotic platforms. Wang \textit{et~al}.~\cite{wang2025} investigate Large Model (LM) agents and address technologies, cooperation mechanisms and challenges related to privacy and security in collaborative environments.

While LLMs are used in many domains of UAVs and wireless systems, there is still a large gap in using LLM-driven flight resource allocation to minimize average AoI in UAWM. To address this gap, we propose FRSICL, a novel framework that utilizes ICL for real-time UAV decision making. FRSICL optimizes the UAV's velocity and data collection schedules over a finite horizon using natural language task prompts derived from sensor logs and continuously adapts via performance feedback to refine decisions.

\section{System Model} \label{sec4}

The proposed UAWM architecture consists of $N$ terrestrial sensor nodes and an autonomous UAV platform. The UAV follows a predefined trajectory marked with multiple waypoints to ensure complete coverage of all sensor deployment areas. This fixed-path strategy reduces computational complexity and energy consumption compared to real-time path planning methods. The FRSICL system is designed to be trajectory independent and to maintain its functionality across different flight patterns. The position of the UAV at time $t$ is represented by the coordinate vector $\zeta(t)$. Its primary objective is to collect and aggregate environmental measurements from each ground sensor $j$ with $j \in \{1, \ldots, N\}$.
\par
The coordinates $(x, y, h)$ and $(x_j, y_j, 0)$ represent the positions of the UAV and ground sensor \emph{$j$} respectively. The UAV flies over the sensor field at optimized velocities, collects sensor data, and makes real-time decisions using the proposed FRSICL system. To ensure safe operation during flight and prevent both overspeed and stall, we define the maximum and minimum allowable velocities of the UAV as $v_{\text{max}}$ and $v_{\text{min}}$, respectively. In this work, the minimum velocity $v_{\text{min}}$ is set to zero.
\par
The UAV operates at low-altitude flight paths to facilitate data acquisition, with the LoS connection probability between the aerial platform and ground sensor $j$ expressed in Eq.~\ref{eq:2} where \emph{$a$} and \emph{$b$} are environment-dependent constants, and  $\varphi_j$ denotes the elevation angle between the UAV and ground sensor \emph{$j$}\cite{al2014optimal}. 

\begin{equation} \label{eq:2}
    \Pr{_{LoS}}(\varphi_j)=\frac{1}{1+a \exp(-b[\varphi_j-a])}
\end{equation}

The elevation angle of ground sensor $j$ is given by Eq.~\ref{eq:20}, where \emph{$h$} is the height of the UAV above the ground and $d_j = \sqrt{(x - x_j)^2 + (y - y_j)^2}$ is the horizontal distance between the UAV and ground sensor $j$.

\begin{equation} \label{eq:20}
\varphi_j = \tan^{-1}\left(\frac{h}{d_j}\right),
\end{equation}

The path loss of the channel between the UAV and ground sensor $j$ is given by Eq.~\ref{eq:3} where $r$ is the radius of the radio coverage of the UAV, $\lambda$ is the carrier frequency, and $v_c$ is the speed of light. $\eta_{LoS}$ and $\eta_{NLoS}$ are the excess path losses of LoS or non-LoS, respectively~\cite{emami2021joint}.
\begin{multline} \label{eq:3}
  \gamma_j=\Pr{_{LoS}}(\varphi_j)(\eta_{LoS}-\eta_{NLoS})+20 \log \left(r \sec(\varphi_j)\right)+ \\20 \log(\lambda)+20 \log \left(\frac{4\pi}{v_c}\right)+\eta_{NLoS}
\end{multline}

\begin{figure} [t]
    \centering 
    \captionsetup{justification=raggedright}
    \includegraphics[width=0.8\columnwidth]{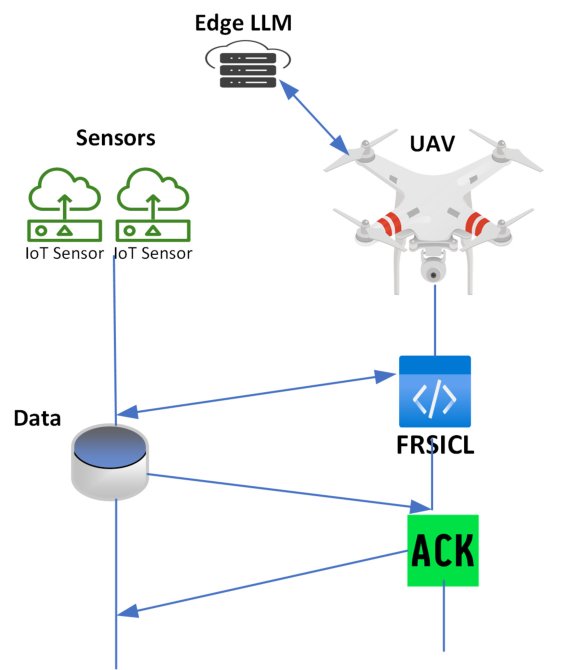}
    \caption{Data communication protocol for the proposed FRSICL, including prompting the LLM, UAV movement and sensor selection, sensor transmission, acknowledgement and update, and iteration.}
    \label{fig:digital40}
\end{figure}

\subsection{Communication Protocol}

The proposed communication protocol for data collection in UAWM using the FRSICL framework is shown in Fig.~\ref{fig:digital40}. 

 In our system, the UAV interacts with the edge-hosted LLM at time step $t$ to determine its flight velocity $v(t)$ and the data collection schedule $\mathcal{A}(t)$. The decision process is based on three key inputs: the Age of Information $AoI_j(t)$ of each sensor $j$, the channel conditions $\gamma_j$, and the current UAV location $\zeta(t)$. The communication and execution procedure unfolds as follows:

\begin{enumerate}
    \item \textbf{Prompting the LLM:} At the beginning of each time step $t$, the UAV sends a query to the edge-hosted LLM, providing the current system state $(AoI_j(t), \gamma_j, \zeta(t))$. The LLM processes this information and outputs the scheduling decision, i.e., the next sensor to be served and the UAV velocity for data collection.
    
    \item \textbf{UAV movement and sensor selection:} Following the LLM’s recommendation, the UAV moves towards the selected sensor at the designated velocity. Once at an appropriate communication distance, the UAV broadcasts a beacon message containing the sensor ID to initiate the data collection process.
    
    \item \textbf{Sensor transmission:} Upon receiving the beacon, the selected sensor responds by transmitting its data packets. These packets include both the sensor readings and auxiliary status information (such as its $AoI_j$ and channel quality indicators).
    
    \item \textbf{Acknowledgement and update:} After successfully receiving the data packets, the UAV verifies their integrity and sends an acknowledgement message back to the sensor. This confirms successful delivery and closes the communication loop with the sensor.
    
    \item \textbf{Iteration:} The UAV then proceeds to the next scheduled sensor. After a predefined interval, a new time step begins, and the UAV again queries the LLM for an updated schedule. This iterative prompting ensures that the UAV’s trajectory and data collection remain adaptive to time-varying conditions.
\end{enumerate}    

Here the UAV acts as the mobile collector and executor, the sensors serve as data sources, and the edge-hosted LLM provides real-time scheduling intelligence. Together, these interactions enable an adaptive and efficient UAV-assisted data collection process.

\subsection{Problem Formulation}

To evaluate the temporal relevance (\textit{i.e.}, data freshness) of the recorded sensor data, we use the Age of Information (AoI) metric. Formally, AoI for sensor $j$ quantifies the time interval between the current observation time $t$ and the creation timestamp of the last successfully collected data sample $U_j(t)$ as shown in Eq.~\ref{eq:aoi}.% denotes the generation time of the last collected packet from ground sensor $j$ before time $t$. Accordingly, the AoI of sensor $j$ at time $t$ is defined as

\begin{equation}
 \mathrm{AoI}_j(t) = t - U_j(t)
 \label{eq:aoi}
\end{equation}

%\color{blue} The AoI for a sensor \(j\) is a metric for data freshness, calculated as the simple difference between the current time \(t\) and the timestamp \(U_j(t)\) of the most recent measurement packet successfully received from that sensor.
 %A smaller value of $\mathrm{AoI}_j(t)$ corresponds to fresher sensor information. 
In UAWM, $\mathrm{AoI}_j(t)$ is significantly influenced by the path loss. Lower path loss means lower signal attenuation between the ground sensors and the UAV, resulting in stronger received signals, lower bit error rates, fewer retransmissions and faster data transmission. This improved communication efficiency implies that the timestamps $U_j(t)$ will be more recent when collected at time $t$, directly lowering $\mathrm{AoI}_j(t)$ and maintaining the freshness and relevance of sensor data for real-time applications. In contrast, high path loss leads to weaker signal reception, more transmission errors and more frequent retransmissions, which increases $\mathrm{AoI}_j(t)$ affecting the timeliness and reliability of the captured data. Other factors that strongly influence $\mathrm{AoI}_j(t)$ in UAWM are the UAV velocity $v(t)$ and the data collection schedule $\mathcal{A}(t)$. While the UAV velocity affects travel time and channel conditions, where higher velocities may increase path loss and transmission delays, the data collection schedule dictates how frequently each sensor is served. %Efficient joint optimization of $v(t)$ and $\mathcal{A}(t)$ is therefore essential for minimizing AoI and ensuring timely information collection.
\par
Therefore, we set the objective of jointly optimizing the UAV velocity $v(t)$ and data collection schedule $\mathcal{A}(t)$ to minimize $\mathrm{AoI}_j(t)$ averaged over all $N$ sensors over a finite horizon that corresponds to the collection schedule as expressed in Eq.~\ref{eq:optim}. 

\begin{equation}
    \min_{\{v(t), \mathcal{A}(t)\}} \quad \frac{1}{N} \sum_{j=1}^{N} \mathrm{AoI}_j(t),
    \label{eq:optim}
\end{equation}
%The function to be minimized is the average AoI across all \(N\) ground sensors at time \(t\). This is calculated by summing the individual AoI value \(\mathrm{AoI}_j(t)\) for each sensor \(j\) and then dividing by the total number of sensors \(N\). 
which defines the core optimization problem: finding the best combination of velocity and data collection schedule that results in the freshest possible data from the entire sensor network. 
%The relationship between UAV velocity and AoI in our formulation is implicit and arises from the wireless channel conditions. Swift UAV movements result in poor channel conditions and rapid signal attenuation, leading to increased AoI. However, slow UAV movements extend flight time, thereby also increasing the sensors data AoI. 
It is worth noting that UAV velocity affects AoI through two coupled mechanisms: (i) mobility efficiency (travel time) and (ii) wireless reliability (path loss, packet errors, and retransmissions). Therefore, AoI does not necessarily decrease monotonically with velocity in UAWM. This coupling also implies that overly aggressive velocities may waste both propulsion effort and transmission opportunities, whereas overly conservative velocities prolong flight time and delay updates.
\par
This work demonstrates the effectiveness of the proposed joint optimization framework in a single UAV scenario, providing clear insights into AoI minimization. %Although this study focuses on a single UAV to highlight the core concepts, the framework can be naturally extended to multi-UAV systems. For example, several UAVs can compete for the sensors, each with its own schedule and speeding up data collection, or several UAVs can be spread over the same data collection path increasing collection frequency. Multi-UAV systems are left for future work. \color{black}
However, the proposed model is not inherently limited to a single UAV and can be naturally extended for multiple UAVs. For example, with multiple UAVs each UAV can operate a subset of sensors and maintain the respective AoI metrics.
The optimization problem can be extended by introducing a data collection schedule and velocity
control for each UAV to minimize the global average AoI while enforcing collision-free routes. We
believe this is a promising direction for future research and emphasize that our results
for a single UAV provide valuable insights that can guide the design of efficient multi-UAV
coordination strategies.

\section{Proposed FRSICL} \label{sec5}

%This section proposes and presents  FRSICL, a method is presented. 
ICL enables LLMs to improve their performance on specific tasks by utilizing structured natural language input, such as task descriptions and solution demonstrations. This process can be formally represented by Eq.~\ref{eq:100} where \( TD_{\text{task}} \) denotes the task description, $Ex_t$ represents the set of examples at time $t$, $e_t$ corresponds to the environmental state associated with the target task at time $t$, \( \text{LLM} \) refers to the large language model, and $a_t$ denotes the model output.     

\begin{equation} \label{eq:100}
TD_{\text{task}} \times Ex_t \times e_t \times \text{LLM} \Rightarrow a_t,
\end{equation}

For sequential decision-making problems, the LLM is expected to process the initial task description \( TD_{\text{task}} \), extract feedback from the example set $Ex_t$, and generate decisions $a_t$ based on the current environment state $e_t$.
The task descriptor $TD_{\text{task}}$ plays a pivotal role in contextualizing the LLM operation by formally specifying:

\begin{itemize}
    \item \textbf{Objective}: The target goals and success criteria
    \item \textbf{Input Schema}: Structure and semantics of input data
    \item \textbf{Operational Constraints}: Rules and limitations governing task execution
    \item \textbf{Output Requirements}: Expected format and properties of generated solutions
    \item \textbf{Feedback Mechanism}: Iterative refinement process for performance improvement
\end{itemize}
This task description avoids the complexity of dedicated optimization model design and relieves the operator from expert knowledge on optimization techniques.

\begin{figure*}[t]
    \centering
    % First row of figures
    \begin{minipage}{0.48\textwidth}
        \centering
        \includegraphics[width=\textwidth]{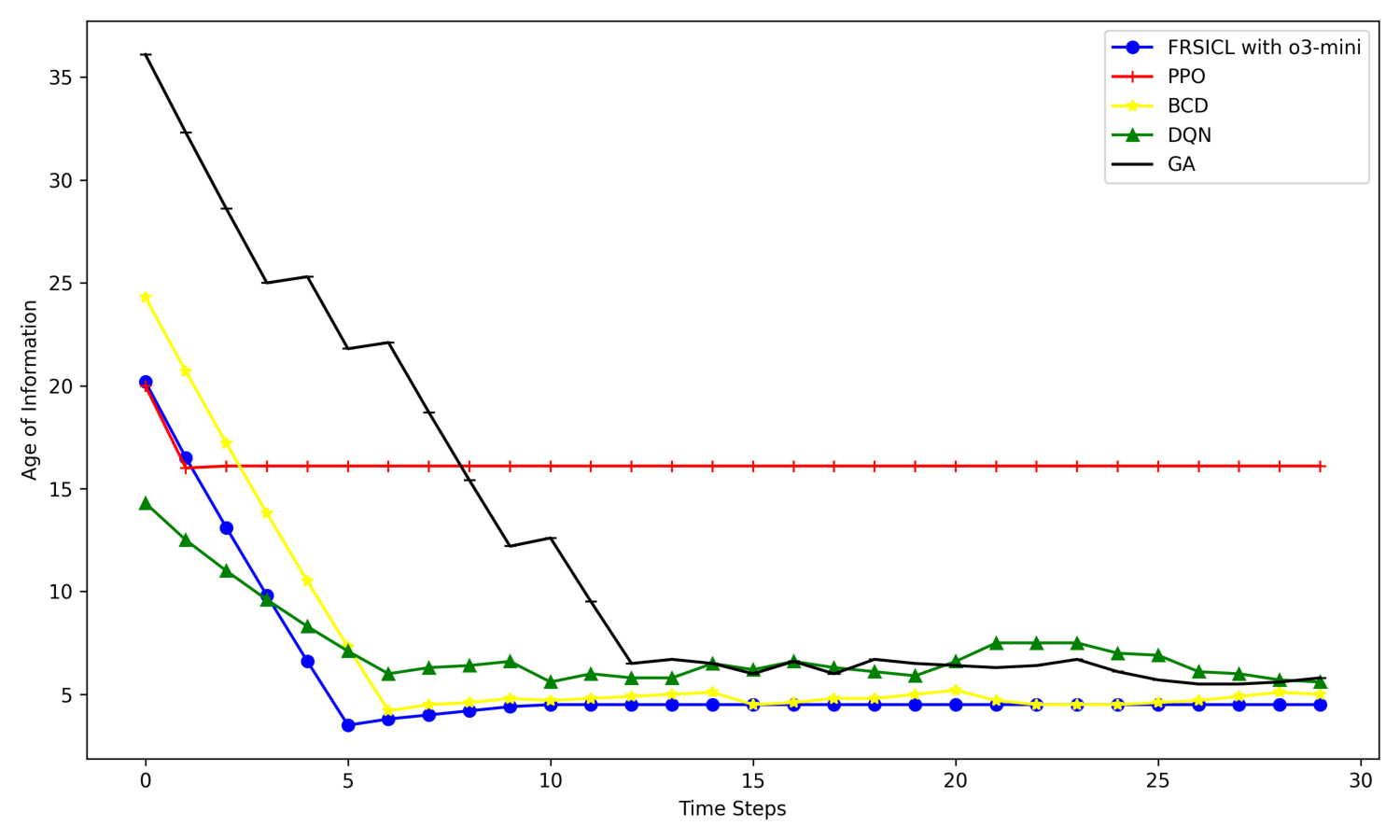}
        \\ \textbf{(a)}
    \end{minipage} \hfill
    \begin{minipage}{0.48\textwidth}
        \centering
        \includegraphics[width=\textwidth]{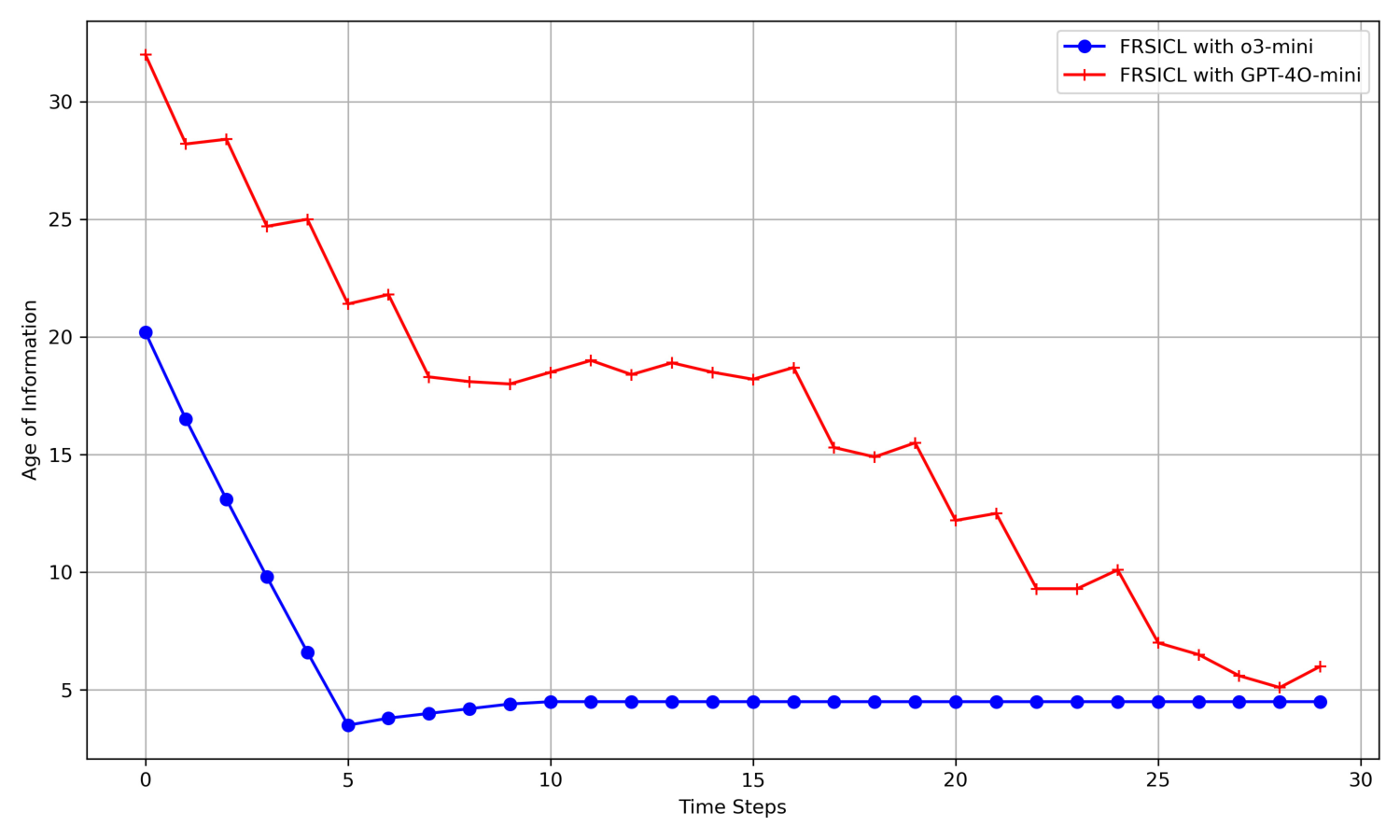}
        \\ \textbf{(b)}
    \end{minipage}
    
    \vspace{1em} % Space between rows

    % Second row of figures
    \begin{minipage}{0.45\textwidth}
        \centering
        \includegraphics[width=\textwidth]{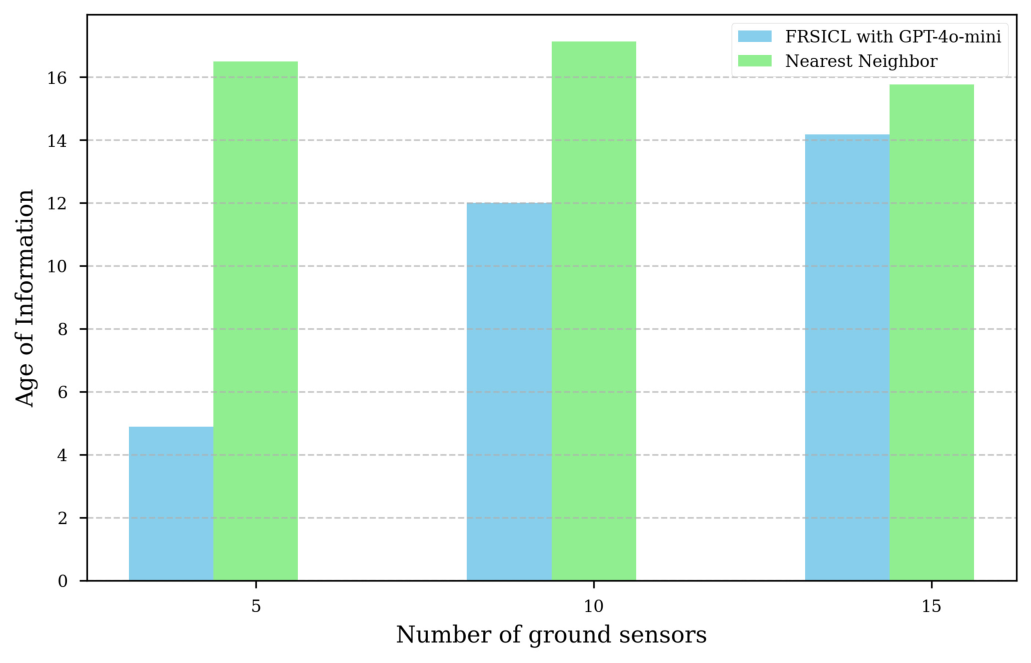}
        \\ \textbf{(c)}
    \end{minipage} \hfill
    \begin{minipage}{0.5\textwidth}
        \centering
        \includegraphics[width=\textwidth]{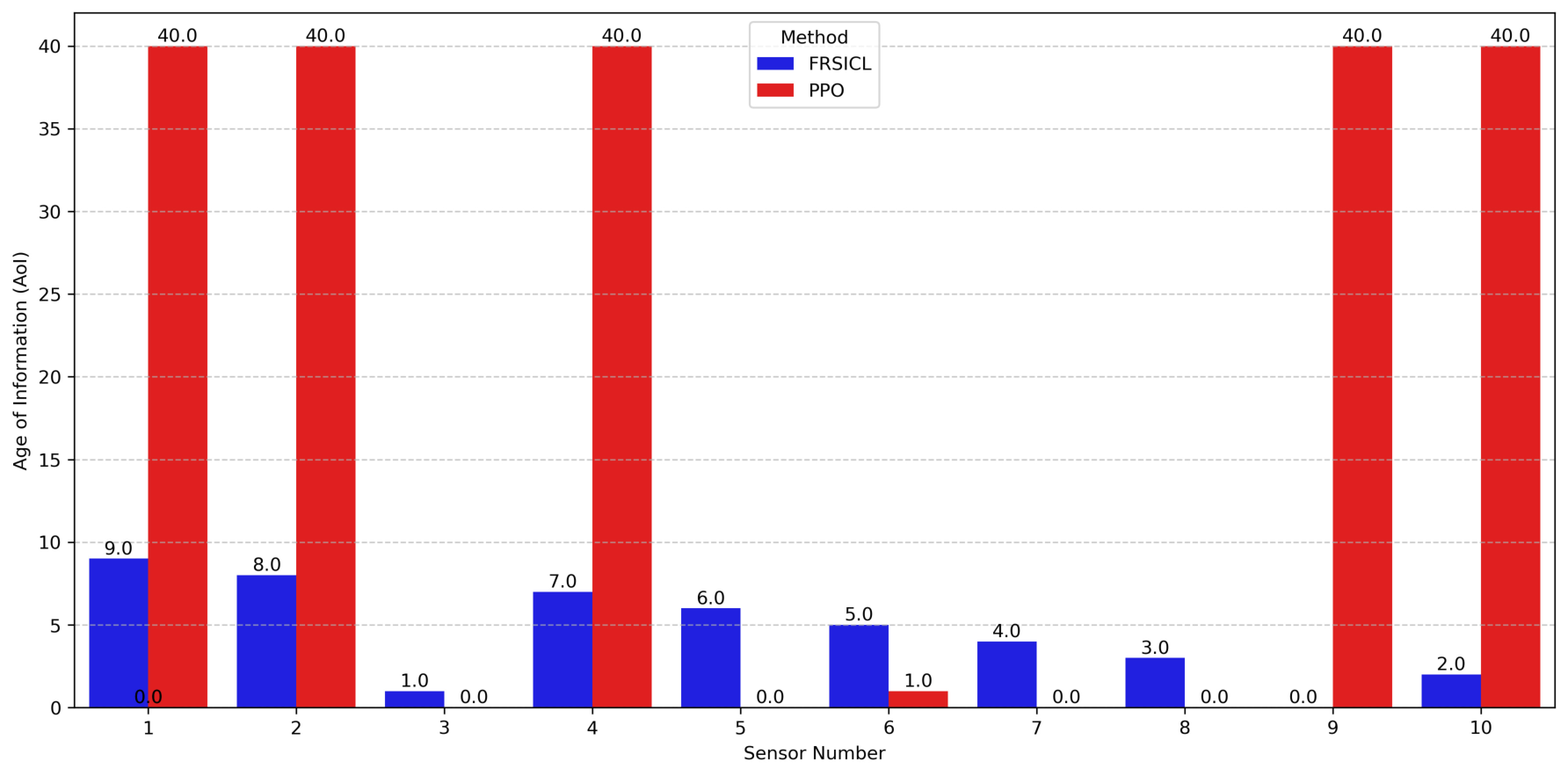}
        \\ \textbf{(d)}
    \end{minipage}

    % Single caption for all figures
    \caption{Performance analysis of FRSICL in different scenarios with 10 ground sensors \textbf{(a) Performance of FRSICL against BCD, and PPO baselines over time steps}} (b) The network cost at each time step of FRSICL with different LLMs (c) Evaluation of FRSICL versus nearest-neighbor approach with different number of sensor nodes (d) AoI comparison between FRSICL and PPO over sensor nodes.
    \label{fig:common_legend}
\end{figure*}

\begin{algorithm}[t]
\caption{Proposed FRSICL: Flight Resource Allocation Using ICL }
\begin{algorithmic}[1]
  \REQUIRE System prompt for LLM, initialized UAV, edge server, sensor nodes, $\mathit{trajectory\_log}$, \; $\mathit{data\_log}$  
  \WHILE{mission is active}
    \STATE \textbf{Sense environment:}
    \STATE \quad $\mathit{sensor\_data} \gets \text{UAV.collect\_data(sensor\_nodes)}$
    
    \STATE \textbf{Transmit to edge:}
    \STATE \quad $\text{edge\_server.receive}(\mathit{sensor\_data})$
    
    \STATE \textbf{Prepare LLM input:}
    \STATE \quad $\mathit{features} \gets \{\mathit{Age\_of\_Information},$  
    \STATE \quad \quad $\mathit{channel\_conditions}, \mathit{UAV\_Location}\}$  
    \STATE \quad {$\mathit{safety\_constraints} \gets \{\mathit{velocity\_bounds}$, 
    \STATE $\mathit{channel\_awareness}, \mathit{diversity\_rules}\}$}
    
    \STATE \textbf{Generate Data Collection schedule:}
    \STATE $\mathit{Schedule} \gets \text{LLM.generate}(\mathit{features})$

    \STATE  $\mathit{safety\_constraints}$)
    
    \STATE \textbf{Generate Velocity:}
    \STATE \quad $\mathit{Velocity} \gets$ \text{LLM.generate}($\mathit{features},$ 
    \STATE $\mathit{safety\_constraints})$
    
    \STATE \textbf{Execute schedule and velocity:}
    \STATE \quad $\text{UAV.execute\_schedule}(\mathit{schedule}, \mathit{velocity})$
    
    \STATE \textbf{Collect feedback:}
    \STATE \quad $\mathit{metrics} \gets \text{UAV.collect\_feedback}() $, e.g., AoI evolution, transmission success, and safety-related outcomes
    
    \STATE \textbf{Update LLM with feedback:}
    \STATE \quad $\text{edge\_server.update\_model}(\mathit{metrics})$
    
    \STATE \textbf{Log trajectory and data:}
    \STATE \quad $\mathit{trajectory\_log}.\text{append}(\text{UAV.current\_path}())$
    \STATE \quad $\mathit{data\_log}.\text{append}(\mathit{sensor\_data})$
   \ENDWHILE
  \ENSURE Final outputs:  
  \STATE \quad UAV trajectory log: $\mathit{trajectory\_log}$
  \STATE \quad Collected sensor data: $\mathit{data\_log}$
\end{algorithmic}
\label{alg:ICL_data_scheduling}
\end{algorithm}

\subsection{Introducing FRSICL}

Algorithm~\ref{alg:ICL_data_scheduling} presents the proposed FRSICL scheme, where LLM-enabled In-Context Learning (ICL) is employed to minimize the average AoI with the following details:
\begin{enumerate}
    \item \textbf{Data Collection}: The UAV collects sensory data, including AoI, channel condition, and the UAV location.
    
    \item \textbf{Contextual Understanding}: The LLM analyzes the collected sensory data according to its trained model. 
    At this stage, safety awareness is implicitly established by enabling the LLM to reason over physical (UAV motion), communication (channel quality), and information freshness states in a unified manner. This contextual understanding allows the LLM to identify potentially unsafe operating conditions, such as excessive velocity under poor channel quality, before making scheduling or control decisions.
    
    \item \textbf{Data Collection Scheduling}: LLM-enabled ICL analyzes the gathered data and schedules data collection and control velocity to minimize the average AoI. 
    Safety-aware prompting is explicitly enforced in this step by embedding operational safety constraints into the scheduling logic. These include (i) velocity-constrained decision making to ensure stable UAV motion and reliable communication, (ii) channel-aware sensor selection to avoid transmission failures caused by poor link conditions, and (iii) diversity-preserving scheduling rules that prevent repeated selection of the same sensor and mitigate information starvation across the network.
    Therefore, the LLM is restricted to data collection tasks and cannot be coerced into adopting unrelated or harmful roles. 
    
    \item \textbf{Adaptive Learning and System Update}: The LLM continuously updates its flight resource allocation based on new data received from the ground sensors. The LLM records its decisions, environment state, and system performance to derive feedback. 
    This feedback-driven update process further reinforces safety by allowing the LLM to learn from unsafe or inefficient past decisions, such as failed transmissions or unstable velocity selections, and adjust future scheduling and control actions accordingly.
    When the next environment state arises, relevant feedback is retrieved to guide subsequent decisions. 
\end{enumerate}

FRSICL uses text input for efficient communication. Token reduction techniques remove redundant input tokens so that the quality of inference is maintained while reducing the transmission payload. This reduces the energy consumption of the UAV and extends that battery life and mission duration~\cite{10835069}.

FRSICL is vulnerable to blackbox jailbreaking attacks, in which adversaries craft malicious prompts, \textit{e.g.}, using techniques such as injection, obfuscation, or replay, to bypass security filters and generate malicious output without access to the model internal parameters. These attacks pose a significant risk, including the spread of misinformation, the generation of unethical content and potential data loss. Mitigation strategies include Reinforcement Learning from human feedback (RLHF), adversarial training and output moderation. However, complete prevention of jailbreaks remains an open challenge as techniques for exploiting jailbreaks continue to evolve~\cite{emami2025ll}.

\begin{table}[t]

\centering
\caption{LLM configuration and deployment details for reproducibility.}
\renewcommand{\arraystretch}{1.15}
\begin{tabular}{m{2.7cm} p{5.7cm}}
\hline
\textbf{Item} & \textbf{Details} \\
\hline
LLM models & o3-mini (2025), GPT-4o-mini \\
\hline
Model type & Compact reasoning models with strong STEM/coding abilities \\
\hline
Model modifications & None (no pruning, quantization, distillation, or fine-tuning) \\
\hline
Deployment location & Edge server (LLM not executed on UAV) \\
\hline
UAV compute overhead & Zero (UAV only sends telemetry/control packets) \\
\hline
Inference method & Remote API (low-latency HTTP/gRPC) \\
\hline
Motivation & High reasoning efficiency with minimal latency and cost \\
\hline
\end{tabular}
\label{tab:llm_deployment}
\end{table}

\begin{table}[t!]
\centering
\renewcommand{\arraystretch}{1.15}
\caption{Configuration}
\begin{tabular}{|m{4 cm}<{\centering}||m{2.5cm}<{\centering}|} 
 \hline
 \textbf{Parameters} & \textbf{Values} \\  
 \hline
 Number of ground sensors ($N$)& 10  \\ 
 \hline
  Queue length ($D$)& 40 \\
 \hline
 Number of Time Steps & 30 \\
 \hline
  Maximum transmit power & 100 mW\\
 \hline
 Maximum velocity & 15 m/s\\
 \hline
 \end{tabular}
\label{table:2}
\end{table}

\section{Numerical Results and Discussions} \label{sec6}

This section presents network configurations and evaluates the network performance of FRSICL.

\subsection{Implementation of FRSICL}

We consider a scenario where 10 ground sensors are randomly placed in a 100 m × 100 m area according to a uniform distribution. The simulation is implemented in Python 3.10. The actions of the UAVs are controlled by policies learned through the proposed FRSICL framework. The main simulation parameters are summarized in Table~\ref{table:2}. Each ground sensor is modeled with a maximum battery capacity of 50 J, a maximum transmit queue length of 40, and a maximum transmit power of 100 mW. Each episode consists of 30 time steps. The wireless communication between the UAV and the ground sensors is simulated using the probabilistic channel model described in Section~\ref{sec4}. All experiments are run on a Lenovo workstation running Ubuntu 20.04 LTS, equipped with an Intel Core i5-7200U CPU @ 2.50 GHz × 4 and 16 GiB RAM.

\subsection{Baselines}

To evaluate the performance of the proposed FRSICL, we compare it with several representative baselines commonly adopted in UAV-assisted data collection and scheduling problems.
1) \textbf{Proximal Policy Optimization (PPO):} PPO is implemented as a learning-based baseline, where a UAV with an actor--critic architecture jointly selects sensors and controls its velocity to minimize the average AoI.
2) \textbf{Block Coordinate Descent (BCD):} BCD is implemented as an iterative optimization baseline, where sensor scheduling and velocity control are alternately optimized to reduce the average AoI.
3) \textbf{Nearest Neighbor (NN):} The Nearest Neighbor baseline selects the closest sensor within communication range at each time step, without explicitly considering AoI dynamics or channel conditions.
4) \textbf{Genetic Algorithm (GA):} GA is included as a heuristic optimization baseline that searches for scheduling and velocity solutions through evolutionary operations without relying on gradient information. 5) \textbf{Deep Q-Network (DQN):} DQN is considered as a lightweight learning-based baseline, where the UAV learns a value function to guide joint sensor selection and velocity control decisions.

\subsection{Performance Analysis}

%Fig.~\ref{fig:common_legend}(a) compares the performance of the proposed FRSICL with the PPO and BCD baselines and illustrates the behavior of the converged policy. All curves start with high AoI values (about 20), reflecting initial data inconsistencies, but diverge as the simulation progresses. PPO shows slower convergence and maintains a relatively high and stable AoI (16–18) until step 30, indicating suboptimal performance. In contrast, the proposed FRSICL converges faster and stabilizes at a significantly lower AoI (5–6). BCD starts with a higher AoI and converges later than FRSICL; however, it stabilizes at a similar AoI. This demonstrates the effectiveness of FRSICL in acquiring fresher data through joint optimization of data collection schedule and velocity control.

Fig.~\ref{fig:common_legend}(a) compares the performance of the proposed FRSICL with the PPO, BCD, DQN, and GA baselines. All methods start with relatively high AoI values due to initial data inconsistencies, but their behaviors diverge over time. PPO exhibits slow convergence and maintains a relatively high AoI throughout the simulation. 
The DQN and GA baselines achieve gradual AoI reduction but converge more slowly and exhibit higher variability compared to FRSICL. BCD shows improved performance but still converges later than FRSICL. In contrast, the proposed FRSICL rapidly converges and stabilizes at a significantly lower AoI level (around 5--6), demonstrating its effectiveness in jointly optimizing data collection scheduling and UAV velocity control.

Fig. \ref{fig:common_legend} (b) compares the performance of the proposed FRSICL with o3-mini with that of GPT-4O-mini. FRSICL with o3-mini achieves faster convergence and lower AoI in comparison to that of GPT-4O-mini, since o3-mini generally outperforms GPT-4o-mini in reasoning and coding tasks especially in complex scenarios like UAWM.  
\par
Fig.~\ref{fig:common_legend}(c) shows a comparison of the AoI as a function of the number of ground sensors for two methods: FRSICL with GPT-4O-mini (shown in sky blue) and the Nearest Neighbor baseline (shown in light green). The x-axis represents the number of ground sensors used (5, 10 and 15), while the y-axis shows the corresponding AoI values. As the number of sensors increases, the AoI value increases for both methods, as it becomes increasingly complicated to obtain fresh data on more nodes. However, FRSICL with GPT-4O-mini consistently achieves a lower AoI than the basic variant. In particular, with 5 sensors, FRSICL maintains an AoI of about 4.9, compared to 16.5 for Nearest Neighbor. Even with 15 sensors, FRSICL maintains a lower AoI (~14.2 vs. ~15.7), demonstrating superior scalability and effectiveness in managing information freshness. These results emphasize the benefits of integrating GPT-4O-mini into the FRSICL system, enabling more efficient scheduling and decision-making in distributed sensor environments.
\par
Fig.~\ref{fig:common_legend}(d) shows a comparative analysis of AoI performance between the proposed FRSICL and the PPO baseline across ten ground sensors in a UAWM scenario. The x-axis represents the sensor indices (1–10), while the y-axis shows the AoI values in seconds, reflecting the timeliness of the collected data. FRSICL (shown in blue) consistently outperforms PPO (red) by maintaining significantly lower AoI values (typically below 10 seconds - while PPO often ends at 40 seconds). The most notable improvement is observed at sensor 5, where FRSICL achieves an AoI value of just 1.0 second compared to 40.0 seconds for PPO, demonstrating FRSICL’s superior adaptability in dynamic environments. This result underscores the effectiveness of FRSICL in ensuring timely data collection, which is a critical requirement for emergency response.

\begin{figure}[t]
    \centering 
    \captionsetup{justification=raggedright}
    \includegraphics[width=0.48\textwidth]{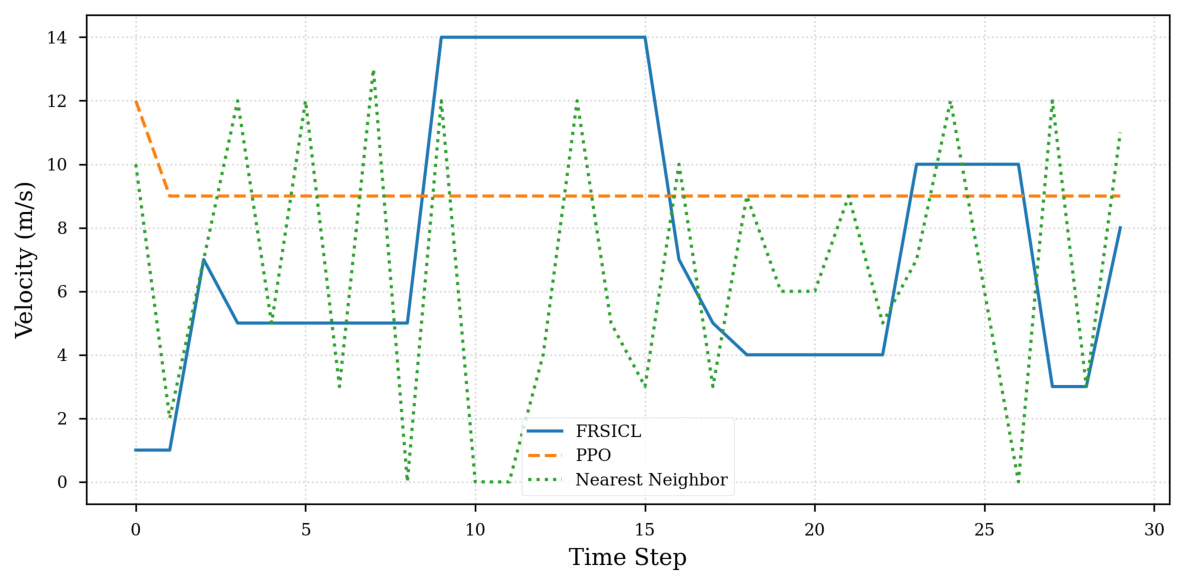}
    \caption{Comparative velocity trajectories during UAV-assisted wildfire monitoring. FRSICL achieves lower average velocity than PPO and smoother velocity than Nearest Neighbor.}
    \label{fig:digital2300}
\end{figure}

\begin{figure}[t]               
    \centering                  
    \includegraphics[width=0.4\textwidth]{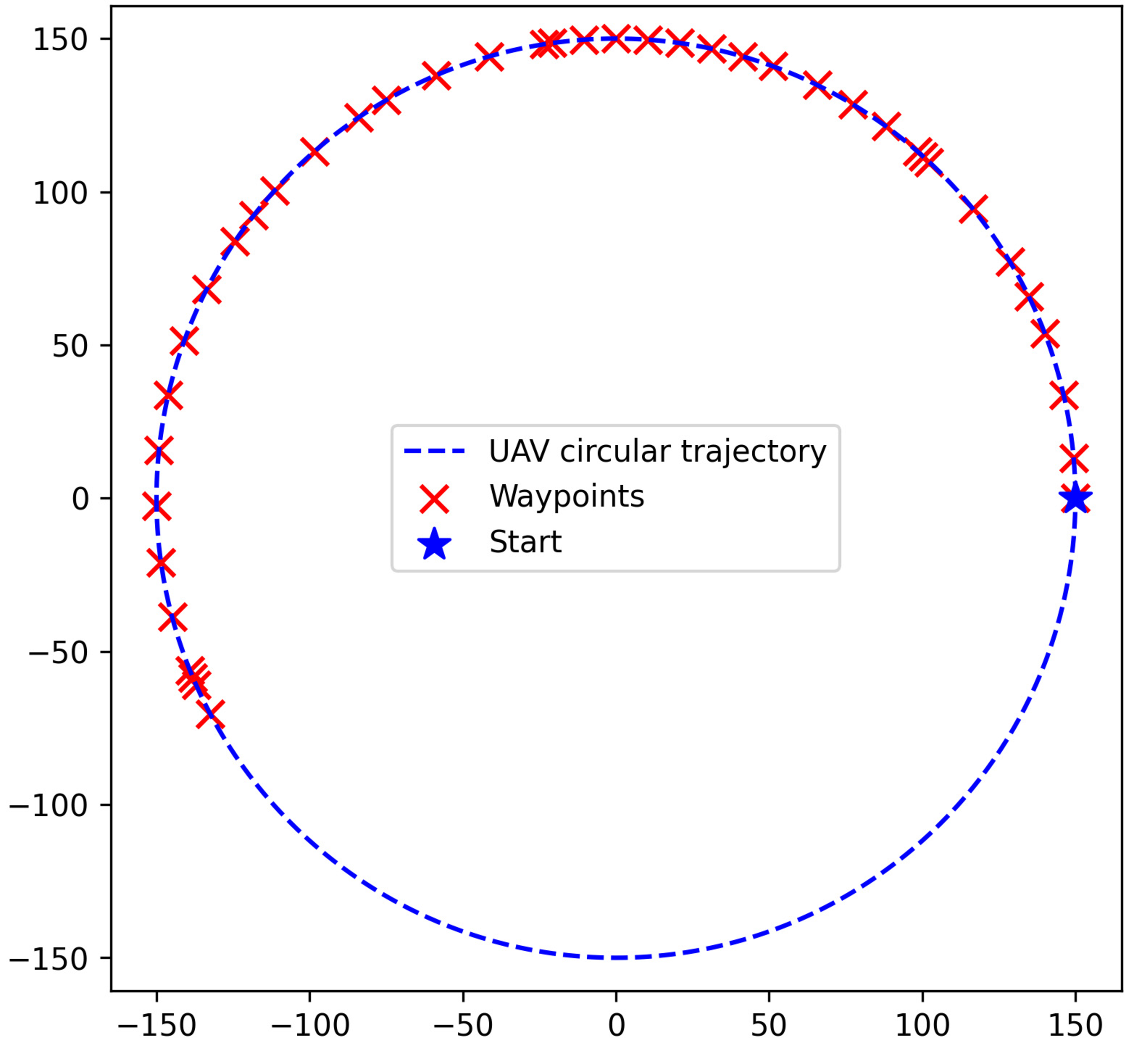}
    \caption{\textbf{UAV circular trajectory with selected waypoints}}
    \label{fig:myphoto}
\end{figure}

Fig.~\ref{fig:digital2300} compares the UAV velocity profiles of three methods in the UAWM environment: the proposed FRSICL approach, PPO and Nearest Neighbor. The x-axis represents time steps, while the y-axis shows the UAV velocity (in m/s). FRSICL (blue solid line) maintains stable velocities and avoids abrupt accelerations or decelerations that could interrupt data collection. In contrast, the Nearest Neighbor method (orange dashed line) exhibits frequent velocity spikes, indicating an inefficient and reactive scheduling strategy. The PPO method (green dashed line) shows a more conservative and overly uniform velocity profile, potentially limiting responsiveness to dynamic sensing demands.
Finally, Fig. \ref{fig:myphoto} serves as a visual representation of a circular flight pattern for the UAV. It shows the path the UAV will take by passing a series of waypoints over a circle. In addition, note that velocity impacts AoI via both travel time and transmission success probability, the resulting velocity behavior under FRSICL reflects an implicit trade-off between aggressive motion and reliable update delivery.

\subsection{Discussion}

While conventional DRL methods are effective for general UAWM optimization, they have critical limitations in emergency scenarios. Their dependence on extensive offline training, the complexity of implementation and the known discrepancy between simulation and reality make them unsuitable for time-critical real-world deployments. Although Diffusion Models (DMs) can improve DRL by generating synthetic data to improve generalization, they result in significant computational overhead, limiting their use in environments with limited capabilities. In contrast, the proposed FRSICL framework utilizes LLM-enabled ICL to overcome these challenges. It offers three key advantages for wireless emergency detection: (1) fast, training-free adaptation through natural language prompts; (2) iterative policy refinement through online performance feedback; and (3) transparent, human-interpretable decision-making processes, which are essential for reliability and responsiveness in critical situations such as wildfire monitoring.
\par
The proposed FRSICL framework improves the reproducibility and transparency of AI-driven optimization by replacing the opaque, hyperparameter-dependent nature of DRL with a prompt-based, feedback-guided process. Through structured experimentation using well-documented OpenAI models, the study shows that ICL-based methods can achieve competitive performance while remaining interpretable, verifiable, and easily replicable. Additionally, consistent benchmarking across different OpenAI models highlights the practical trade-offs between reasoning ability, latency, and cost, reinforcing the value of using standardized models as credible baselines for reproducible AI research.

\section{Conclusion} \label{sec7}

UAVs play a critical role in wildfire monitoring, where minimizing AoI through efficient data collection and flight control is essential for timely disaster relief. However, DRL methods often suffer from gaps in simulation and high computational requirements, limiting their practicality in urgent scenarios. To overcome these challenges, this paper presents FRSICL a novel LLM-powered ICL framework that dynamically optimizes UAV velocity and data collection schedule based on natural language task descriptions. By utilizing pre-trained knowledge and environmental feedback, FRSICL eliminates the retraining bottlenecks associated with DRL and enables adaptive online decision making. Simulation results show that FRSICL significantly reduces AoI compared to PPO, BCD and nearest-neighbor baselines while ensuring stable velocity control. Future research will explore the extension to collaborative data collection schedules and velocities control strategies for multiple UAVs to further minimize the average AoI.

\bibliographystyle{IEEEtran}
\bibliography{references}

\end{document}